\documentclass[11pt]{article}
\usepackage[hyperref]{ccl2025-en}
\usepackage{times}
\usepackage{url}
\usepackage{latexsym}
\usepackage{graphicx}
\usepackage{booktabs}
\usepackage{fancyhdr}

\title{System Report for CCL25-Eval Task 10: SRAG-MAV for Fine-Grained Chinese Hate Speech Recognition}

\author{
  Jiahao Wang, Ramen Liu, Longhui Zhang, Jing Li\thanks{Corresponding author: Jing Li (\texttt{jingli.phd@hotmail.com}).} \\
  Harbin Institute of Technology, Shenzhen, China \\
  \texttt{wjh123king@gmail.com, jingli.phd@hotmail.com}
}

\date{}

\begin{document}
\maketitle

\begin{abstract}
This paper presents our system for CCL25-Eval Task 10, addressing Fine-Grained Chinese Hate Speech Recognition (FGCHSR). We propose a novel SRAG-MAV framework that synergistically integrates task reformulation(TR), Self-Retrieval-Augmented Generation (SRAG), and Multi-Round Accumulative Voting (MAV). Our method reformulates the quadruplet extraction task into triplet extraction, uses dynamic retrieval from the training set to create contextual prompts, and applies multi-round inference with voting to improve output stability and performance. Our system, based on the Qwen2.5-7B model, achieves a Hard Score of 26.66, a Soft Score of 48.35, and an Average Score of 37.505 on the STATE ToxiCN dataset, significantly outperforming baselines such as GPT-4o (Average Score 15.63) and fine-tuned Qwen2.5-7B (Average Score 35.365). The code is available at \url{https://github.com/king-wang123/CCL25-SRAG-MAV}.
\end{abstract}

\cclfootnote{
    \textcopyright 2025 China National Conference on Computational Linguistics

    \noindent This work is licensed under a Creative Commons Attribution 4.0 International License. License details: \url{http://creativecommons.org/licenses/by/4.0/}.
}

\section{Introduction}
\label{intro}

The growth of social media has significantly amplified the spread of hate speech \cite{fortuna2018survey}, with malicious content targeting attributes such as race, region, and gender, causing substantial harm to individuals and society. Effective hate speech detection has become a critical focus in Natural Language Processing (NLP), aiming to mitigate these negative impacts \cite{davidson2017automated,waseem2016hateful}. Moreover, ensuring the fairness of detection models to avoid potential biases is essential for their practical deployment \cite{sap2019risk}. Traditional methods often rely on binary classification to identify hateful content \cite{fortuna2018survey}, but these approaches lack the granularity to capture the internal structure of hate speech, limiting their interpretability and utility for downstream applications \cite{yin2021towards}. Consequently, Fine-Grained Chinese Hate Speech Recognition (FGCHSR), which extracts structured information such as specific targets or types of hate, has gained increasing attention \cite{basile2019semeval,mathew2021hatexplain,ren2021table}.

CCL25-Eval Task 10 focuses on extracting quadruplets (Target, Argument, Targeted Group, Hateful) from Chinese social media texts. This task is particularly challenging due to the subtle, context-dependent nature of Chinese hate speech \cite{pavlopoulos2020toxicity}, the interdependence of quadruplet elements, and the limited availability of high-quality annotated data \cite{yin2021towards}. The STATE ToxiCN study \cite{bai2025state} highlights these difficulties, showing that even the most advanced models like GPT-4o achieve an Average Score of only 15.63, while fine-tuned open-source models like Qwen2.5-7B reach 35.365, but still require further optimization.

To address these challenges, we propose a novel SRAG-MAV framework synergistically combining Task Reformulation (TR), Self-Retrieval-Augmented Generation (SRAG), and Multi-Round Accumulative Voting (MAV). Our approach simplifies quadruplet extraction into triplet extraction, enhances contextual understanding through dynamic retrieval inspired by Retrieval-Augmented Generation (RAG) \cite{lewis2020rag}, and ensures stable outputs via multi-round inference based on the principles of Parallel Scaling Law (PARSCALE) \cite{chen2025parscale}. Our contributions include:

\begin{itemize}
    \item Proposed a novel SRAG-MAV framework that integrates TR, SRAG, and MAV, demonstrates superior performance for FGCHSR and is adaptable to other structured NLP tasks.
    \item Conducted comprehensive experiments, validating the effectiveness and robustness of our approach and assessing the performance contributions of individual components.
    \item Released code at \url{https://github.com/king-wang123/CCL25-SRAG-MAV}, promoting reproducibility and facilitating further research in hate speech detection and other related NLP domains.
\end{itemize}

\section{Methodology}
\subsection{System Overview}

\begin{figure}[h]
    \centering
    \includegraphics[width=0.8\textwidth]{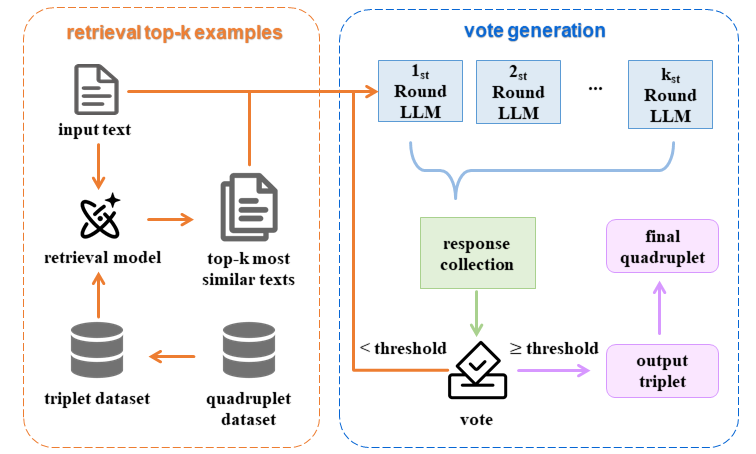}
    \caption{System architecture of SRAG-MAV, depicting the workflow from input text to final quadruplet output. The process includes: (1) transforming the quadruplet dataset into a triplet dataset, (2) retrieving top-\(k\) similar samples and concatenating each with the input text to construct prompts, (3) voting to select triplet answer with frequencies exceeding the threshold \(\tau\), otherwise continuing iterative inference until the threshold is met, and (4) converting the selected triplet into the final quadruplet output.}
    \label{fig:structure}
\end{figure}

Our system combines TR, SRAG, and MAV to extract fine-grained hate speech quadruplets from Chinese social media texts, as shown in Figure~\ref{fig:structure}. The workflow simplifies the task into triplet extraction, enhances contextual understanding through retrieval, and stabilizes outputs via iteratively voting.

Initially, we transform the quadruplet dataset into a triplet dataset to reduce task complexity, aligning with the principles of TR. Then, we encode all training inputs into a vector database using a retrieval model, enabling efficient retrieval for both training and inference phases.

\textbf{Training Phase}: For each input, we retrieve the most similar sample from the training set excluding the input itself, concatenate the retrieved sample with the input to form a prompt. These prompts along with their corresponding triplet outputs, form new training samples for fine-tuning.

\textbf{Inference Phase}: For each test input, we retrieve the top-\(k\) similar training samples to construct \(k\) prompts. The model iteratively generates triplets for these \(k\) prompts across multiple rounds until the frequency of the most frequent triplet exceeds the threshold \(\tau\), at which point MAV selects it as the final triplet answer. This selected triplet is then converted into a quadruplet by inferring hatefulness from the target group.

This pipeline, visualized in Figure~\ref{fig:structure}, balances simplicity, contextual richness \cite{lewis2020rag}, and output stability, with \(k\) and \(\tau\) as key hyperparameters.

\subsection{Task Reformulation (TR)}

Analysis of the training data reveals a strong correlation between the target group and hatefulness label: ``no-hate'' hatefulness occurs only when the target group is ``no-hate''; otherwise, the hatefulness is labeled as ``hate.'' Leveraging this pattern, we reformulate the original quadruplet extraction task into a triplet extraction task. This simplification reduces the complexity of structured generation, as the hatefulness label can be deterministically inferred from the target group, thereby improving the efficiency and accuracy of large language models (LLMs).

\begin{figure}[h]
    \centering
    \includegraphics[width=0.8\textwidth]{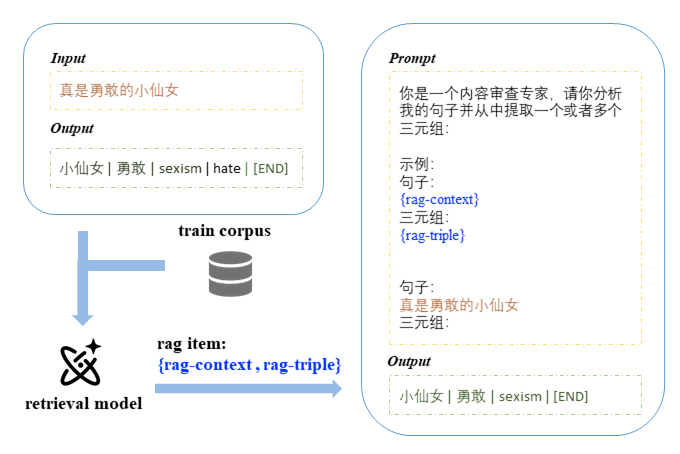}
    \caption{Illustration of TR and SRAG: the retrieval model retrieves similar texts from the training corpus, concatenates them into a prompt, and generates the corresponding triplet.}
    \label{fig:data_demo}
\end{figure}

Figure~\ref{fig:data_demo} provides a concrete data example to illustrate how TR simplifies the extraction process while maintaining the integrity of the structured output.

\subsection{Self-Retrieval-Augmented Generation (SRAG)}

Retrieval-Augmented Generation (RAG) has emerged as a powerful paradigm in NLP, widely applied to tasks such as question answering, dialogue systems, and knowledge-intensive text generation \cite{lewis2020rag,gao2023rag}. By integrating external knowledge through retrieval, RAG enhances the contextual relevance and factual accuracy of generated outputs \cite{ram2023incontext}. Recent advancements have further refined RAG to handle structured data and improve robustness in low-resource settings \cite{zhang2023twostage}. However, FGCHSR poses unique challenges, including the lack of high-quality external corpora and the complexity of structured quadruplet generation.

To address these challenges, we propose the Self-Retrieval-Augmented Generation (SRAG) framework, which adapts the RAG paradigm by using the training set itself as the retrieval corpus. SRAG leverages semantically similar annotated examples to guide triplet generation, ensuring contextually relevant outputs without requiring external resources. The SRAG pipeline includes:

\begin{enumerate}
    \item \textbf{Corpus Construction}: We use the bge-large-zh-v1.5 model \cite{xiao2023bge} to generate embeddings for training set texts, building a retrieval corpus based on cosine similarity.
    \item \textbf{Dynamic Retrieval}: For each input text, we retrieve the most similar sample except the input itself during training, while retrieving the top-k most similar samples during inference.
    \item \textbf{Prompt Generation}: Retrieved samples are combined with the input text to create structured prompts, guiding the model to produce task-compliant triplets, exemplified in Figure~\ref{fig:data_demo}.
\end{enumerate}

SRAG innovatively leverages the training set as a dynamic retrieval corpus, enabling few-shot learning through similar annotated examples to enhance task understanding and output accuracy. Unlike traditional RAG, SRAG eliminates the need for external data, making it particularly suited for resource-constrained environments and domain-specific tasks like FGCHSR.

\subsection{Multi-Round Accumulative Voting (MAV)}

Parallel Scaling Law (PARSCALE) \cite{chen2025parscale} illustrates that applying diverse transformations to an input to generate multiple variants, followed by parallel inference and result aggregation with learnable parameters, can significantly enhance LLMs performance. This approach improves robustness and accuracy without requiring model retraining, making it efficient for complex tasks in resource-constrained settings. 

Inspired by PARSCALE's emphasis on parallel processing of diverse inputs, we introduce Multi-Round Accumulative Voting (MAV) as an innovative adaptation for FGCHSR, which generates diverse prompts with SRAG-retrieved examples and selects the optimal triplet output through a voting mechanism. The MAV pipeline includes:

\begin{enumerate}
    \item \textbf{Diverse Prompts}: Through SRAG, retrieve the top-\(k\) most similar samples from the triplet dataset for each input text, and concatenate each retrieved sample with the input to construct \(k\) distinct prompts.
    \item \textbf{Multi-Round Inference}: Iteratively perform inference on each prompt, generating and accumulating the frequencies of triplet results across iterations until the most frequent triplet exceeds the threshold \(\tau\).
    \item \textbf{Voting Mechanism}: Select the triplet output reaching a frequency threshold \(\tau\) and convert it to a quadruplet as the final result.
\end{enumerate}

MAV stands out for its cost-effectiveness, requiring only additional inference-time resources rather than retraining or parameter adjustments. Its flexibility is demonstrated in Section~\ref{sec:results}, where increasing the threshold progressively improves results, allowing dynamic adjustment based on available computational resources. Additionally, its straightforward implementation enhances reliability, making MAV particularly effective under constrained conditions.

\section{Experiments}
\label{sec:experiments}

\subsection{Experimental Setup}

We evaluated our approach on the STATE ToxiCN dataset \cite{bai2025state}, comprising 4,000 training samples and 1,602 test samples, using 4$\times$NVIDIA L40S 40GB GPUs. The base model, Qwen2.5-7B \cite{qwen2024technical}, was trained with the LLaMA-Factory framework \cite{zheng2024llamafactory} and deployed for inference using vLLM \cite{kwon2023vllm}. For retrieval, we employed the bge-large-zh-v1.5 model \cite{xiao2023bge}, with generation parameters configured at a temperature of 0.7 during fine-tuning and 0.1 for MAV inference. MAV configuration utilized a top-\(k\) value of 10 and voting threshold \(\tau\) of 200. The evaluation metrics included:

\begin{itemize}
    \item \textbf{Hard Score}: F1 score for precise quadruplet matches.
    \item \textbf{Soft Score}: F1 score for partial matches, requiring identical target group and hatefulness, with Target and Argument similarity exceeding 50\%.
    \item \textbf{Average Score}: Mean of Hard and Soft Scores.
\end{itemize}

Baselines were drawn from the STATE ToxiCN benchmarks \cite{bai2025state}, and our approach was compared to validate its effectiveness.

\subsection{Experimental Results}
\label{sec:results}

We conducted three experiments, including a model comparison to benchmark our system against baselines, an MAV parameter sensitivity analysis to assess the impact of threshold variations on performance, and ablation studies to evaluate the contribution of each component.

\subsubsection{Model Comparison}

\begin{table}[h]
\centering
\begin{tabular}{lccc}
\toprule
\textbf{Model} & \textbf{Hard Score} & \textbf{Soft Score} & \textbf{Average Score} \\
\midrule
mT5-base & 16.60 & 38.61 & 27.605 \\
Mistral-7B & 23.72 & 45.62 & 34.670 \\
LLaMA3-8B & 24.27 & 46.08 & 35.175 \\
Qwen2.5-7B & 23.70 & 47.03 & 35.365 \\
ShieldLM-14B-Qwen & 23.59 & 45.58 & 34.585 \\
ShieldGemma-9B & 23.49 & 47.14 & 35.315 \\
\midrule
Ours & \textbf{26.66} & \textbf{48.35} & \textbf{37.505} \\
\bottomrule
\end{tabular}
\caption{Performance comparison on the STATE ToxiCN test set. Results for baseline models are directly cited from the STATE ToxiCN paper \cite{bai2025state}, representing vanilla Supervised Fine-Tuning (SFT) outcomes. Our approach significantly outperforms these baselines across all metrics.}
\label{tab:model_comparison}
\end{table}

Table~\ref{tab:model_comparison} demonstrates that our system achieves a Hard Score of 26.66, a Soft Score of 48.35, and an Average Score of 37.505 on the STATE ToxiCN test set, significantly surpassing all baseline models trained with vanilla Supervised Fine-Tuning (SFT). Our approach yields substantial improvements, particularly in the Hard Score, which reflects precise quadruplet matches and indicates robust performance in FGCHSR.

\subsubsection{MAV Parameter Sensitivity Analysis}

\begin{figure}[h]
    \centering
    \includegraphics[width=0.95\textwidth]{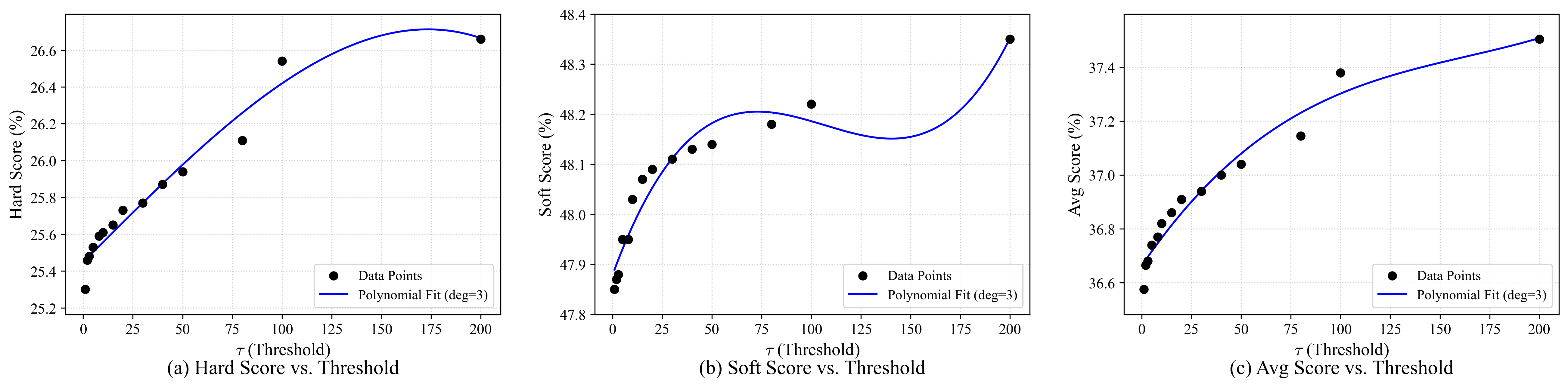}
    \caption{Impact of the MAV threshold parameter (top-\(k\)=10) on the STATE ToxiCN test set. The figure illustrates the relationship between the threshold (\(\tau\)) values and the resulting Hard Score (a), Soft Score (b), and Average Score (c).}
    \label{fig:threshold}
\end{figure}

Figure~\ref{fig:threshold} reveals a detailed analysis of the MAV threshold (\(\tau\)) parameter's impact on model performance, with thresholds tested at [1, 2, 3, 5, 8, 10, 15, 20, 30, 40, 50, 80, 100, 200]. The Hard Score exhibits a pronounced increase from 25.30 to 26.66, reflecting a 1.36-point gain, with notable jumps at higher thresholds (e.g., from 26.11 at \(\tau=80\) to 26.66 at \(\tau=200\)), underscoring the method's effectiveness in achieving precise quadruplet matches. The Soft Score also improves steadily from 47.85 to 48.35, a 0.50-point rise, indicating enhanced partial match accuracy. The Average Score rises from 36.575 to 37.505, a 0.93-point improvement, reflecting a balanced enhancement across both metrics. These results validate MAV's role in stabilizing outputs through accumulative voting, with the Hard Score's significant growth highlighting its superiority in precise fine-grained hate speech recognition.  

Model training provides foundational capabilities, while inference strategies further unleash the model's potential by introducing moderate computational overhead, with the two complementing each other.

\subsubsection{Ablation Study}

\begin{table}[h]
\centering
\begin{tabular}{lccc}
\toprule
\textbf{Configuration} & \textbf{Hard Score} & \textbf{Soft Score} & \textbf{Average Score} \\
\midrule
Base Model & 23.70 & 47.03 & 35.365 \\
+ TR & 24.33 & 47.35 & 35.840 \\
+ TR + SRAG & 25.30 & 47.85 & 36.575 \\
+ TR + SRAG + MAV & \textbf{26.66} & \textbf{48.35} & \textbf{37.505} \\
\bottomrule
\end{tabular}
\caption{Ablation study results, demonstrating the incremental contributions of Task Reformulation (TR), Self-Retrieval-Augmented Generation (SRAG), and Multi-Round Accumulative Voting (MAV) to the overall performance.}
\label{tab:ablation}
\end{table}

As shown in Table~\ref{tab:ablation}, the ablation study offers a high-level perspective on the effectiveness of each component in enhancing the model's performance. Beginning with the base model (Qwen2.5-7B trained via vanilla SFT), the introduction of TR simplifies the model's output structure, leading to a noticeable performance uplift that underscores its effectiveness in streamlining the task. Building on this, the addition of SRAG further strengthens the model by leveraging contextual retrieval, resulting in a clear improvement that highlights its role in refining predictions. The final incorporation of MAV delivers the most significant enhancement, markedly boosting the model's stability and accuracy through iterative inference, which emphasizes MAV's pivotal contribution to overall performance.

\section{Conclusion}

For CCL25-Eval Task 10, we developed a novel SRAG-MAV framework, for the purpose of effectively detecting and mitigating the spread of harmful content on social media. Our approach achieves a Hard Score of 26.66, a Soft Score of 48.35, and an Average Score of 37.505 on the STATE ToxiCN test set \cite{bai2025state}, significantly outperforming baselines such as GPT-4o (Average Score 15.63) and fine-tuned Qwen2.5-7B (Average Score 35.365). TR simplifies the quadruplet extraction task into triplet extraction, reducing complexity; SRAG enhances contextual understanding by leveraging the training set as a retrieval corpus; and MAV ensures output stability through iterative prompt generation and voting. These components work synergistically, as demonstrated by our ablation study, which highlights incremental performance gains from each module.

Our system’s open-source implementation (\url{https://github.com/king-wang123/CCL25-SRAG-MAV}) fosters reproducibility and further research. However, limitations include the model’s domain-specific performance, reliance on text-only data, and MAV’s high voting thresholds increase computational costs. Future work will explore cross-domain transfer learning to enhance generalizability \cite{toraman2022large}, multimodal approaches integrating text and images for richer context \cite{gomez2020exploring,pozo2022detecting}, and optimization of MAV’s computational efficiency to broaden its applicability.

\section*{Acknowledgements}

This work was supported in part by National Science Foundation of China (62476070), Shenzhen Science and Technology Program (JCYJ20241202123503005, GXWD20231128103232001, ZDSYS20230626091203008, KQTD2024072910215406)  and Department of Science and Technology of Guangdong (2024A1515011540).

We thank the CCL25-Eval organizers for their platform, the STATE ToxiCN dataset providers for supporting our experiments, and the reviewers for their valuable feedback.

\bibliography{references}

\begin{thebibliography}{}

\bibitem[\protect\citename{Bai \bgroup et al.\egroup }2025]{bai2025state}
Zewen Bai, Shengdi Yin, Junyu Lu, Jingjie Zeng, Haohao Zhu, Yuanyuan Sun, Liang Yang, and Hongfei Lin.
\newblock 2025.
\newblock State toxicn: A benchmark for span-level target-aware toxicity extraction in chinese hate speech detection.
\newblock {\em arXiv preprint arXiv:2501.15451}.

\bibitem[\protect\citename{Basile \bgroup et al.\egroup }2019]{basile2019semeval}
Valerio Basile, Cristina Bosco, Elisabetta Fersini, Debora Nozza, Viviana Patti, Francisco Manuel~Rangel Pardo, Paolo Rosso, and Manuela Sanguinetti.
\newblock 2019.
\newblock Semeval-2019 task 5: Multilingual detection of hate speech against immigrants and women in twitter.
\newblock In {\em Proceedings of the 13th International Workshop on Semantic Evaluation}, pages 54--63.

\bibitem[\protect\citename{Chen \bgroup et al.\egroup }2025]{chen2025parscale}
Mouxiang Chen, Binyuan Hui, Zeyu Cui, Jiaxi Yang, Dayiheng Liu, Jianling Sun, Junyang Lin, and Zhongxin Liu.
\newblock 2025.
\newblock Parallel scaling law for language models.
\newblock {\em arXiv preprint arXiv:2505.10475}.

\bibitem[\protect\citename{Das \bgroup et al.\egroup }2020]{pozo2022detecting}
Abhishek Das, Japsimar~Singh Wahi, and Siyao Li.
\newblock 2020.
\newblock Detecting hate speech in multimodal memes.
\newblock {\em arXiv preprint arXiv:2012.14891}.

\bibitem[\protect\citename{Davidson \bgroup et al.\egroup }2017]{davidson2017automated}
Thomas Davidson, Dana Warmsley, Michael Macy, and Ingmar Weber.
\newblock 2017.
\newblock Automated hate speech detection and the problem of offensive language.
\newblock In {\em Proceedings of the International AAAI Conference on Web and Social Media}, volume~11, pages 512--515.

\bibitem[\protect\citename{Fortuna and Nunes}2018]{fortuna2018survey}
Paula Fortuna and Sérgio Nunes.
\newblock 2018.
\newblock A survey on automatic detection of hate speech in text.
\newblock {\em ACM Computing Surveys}, 51(4):1--30.

\bibitem[\protect\citename{Gao \bgroup et al.\egroup }2023]{gao2023rag}
Yunfan Gao, Yun Xiong, Xinyu Gao, Kangxiang Jia, Jinliu Pan, Yuxi Bi, Yi~Dai, Jiawei Sun, Meng Wang, and Haofen Wang.
\newblock 2023.
\newblock Retrieval-augmented generation for large language models: A survey.
\newblock {\em arXiv preprint arXiv:2312.10997}.

\bibitem[\protect\citename{Gomez \bgroup et al.\egroup }2020]{gomez2020exploring}
Raul Gomez, Jaume Gibert, Lluis Gomez, and Dimosthenis Karatzas.
\newblock 2020.
\newblock Exploring hate speech detection in multimodal publications.
\newblock In {\em Proceedings of the IEEE/CVF Winter Conference on Applications of Computer Vision}, pages 1470--1478.

\bibitem[\protect\citename{Kwon \bgroup et al.\egroup }2023]{kwon2023vllm}
Woosuk Kwon, Zhuohan Li, Siyuan Zhuang, Ying Sheng, Lianmin Zheng, Cody~Hao Yu, Joseph~E. Gonzalez, Hao Zhang, and Ion Stoica.
\newblock 2023.
\newblock Efficient memory management for large language model serving with pagedattention.
\newblock In {\em Proceedings of the 29th Symposium on Operating Systems Principles}, pages 611--628.

\bibitem[\protect\citename{Lewis \bgroup et al.\egroup }2020]{lewis2020rag}
Patrick Lewis, Ethan Perez, Aleksandra Piktus, Fabio Petroni, Vladimir Karpukhin, Naman Goyal, Heinrich Küttler, Mike Lewis, Wen tau Yih, Tim Rocktäschel, Sebastian Riedel, and Douwe Kiela.
\newblock 2020.
\newblock Retrieval-augmented generation for knowledge-intensive nlp tasks.
\newblock In {\em Advances in Neural Information Processing Systems}, volume~33, pages 9459--9474.

\bibitem[\protect\citename{Mathew \bgroup et al.\egroup }2021]{mathew2021hatexplain}
Binny Mathew, Punyajoy Saha, Seid~Muhie Yimam, Chris Biemann, Pawan Goyal, and Animesh Mukherjee.
\newblock 2021.
\newblock Hatexplain: A benchmark dataset for explainable hate speech detection.
\newblock In {\em Proceedings of the AAAI Conference on Artificial Intelligence}, volume~35, pages 14867--14875.

\bibitem[\protect\citename{Pavlopoulos \bgroup et al.\egroup }2020]{pavlopoulos2020toxicity}
John Pavlopoulos, Jeffrey Sorensen, Lucas Dixon, Nithum Thain, and Ion Androutsopoulos.
\newblock 2020.
\newblock Toxicity detection: Does context really matter?
\newblock In {\em Proceedings of the 58th Annual Meeting of the Association for Computational Linguistics}, pages 4296--4305.

\bibitem[\protect\citename{Ram \bgroup et al.\egroup }2023]{ram2023incontext}
Ori Ram, Yoav Levine, Itay Dalmedigos, Dor Muhlgay, Amnon Shashua, Kevin Leyton-Brown, and Yoav Shoham.
\newblock 2023.
\newblock In-context retrieval-augmented language models.
\newblock {\em Transactions of the Association for Computational Linguistics}, 11:1316--1331.

\bibitem[\protect\citename{Ren \bgroup et al.\egroup }2021]{ren2021table}
Feiliang Ren, Longhui Zhang, Shujuan Yin, Xiaofeng Zhao, Shilei Liu, Bochao Li, and Yaduo Liu.
\newblock 2021.
\newblock A novel global feature-oriented relational triple extraction model based on table filling.
\newblock {\em arXiv preprint arXiv:2109.06705}.

\bibitem[\protect\citename{Sap \bgroup et al.\egroup }2019]{sap2019risk}
Maarten Sap, Dallas Card, Saadia Gabriel, Yejin Choi, and Noah~A. Smith.
\newblock 2019.
\newblock The risk of racial bias in hate speech detection.
\newblock In {\em Proceedings of the 57th Annual Meeting of the Association for Computational Linguistics}, pages 1668--1678.

\bibitem[\protect\citename{Team}2024]{qwen2024technical}
Qwen Team.
\newblock 2024.
\newblock Qwen2.5 technical report.
\newblock {\em arXiv preprint arXiv:2412.15115}.

\bibitem[\protect\citename{Toraman \bgroup et al.\egroup }2022]{toraman2022large}
Cagri Toraman, Furkan Şahinuç, and Eyup Yilmaz.
\newblock 2022.
\newblock Large-scale hate speech detection with cross-domain transfer.
\newblock In {\em Proceedings of the 13th Language Resources and Evaluation Conference}, pages 2215--2225.

\bibitem[\protect\citename{Waseem and Hovy}2016]{waseem2016hateful}
Zeerak Waseem and Dirk Hovy.
\newblock 2016.
\newblock Hateful symbols or hateful people? predictive features for hate speech detection on twitter.
\newblock In {\em Proceedings of the NAACL Student Research Workshop}, pages 88--93.

\bibitem[\protect\citename{Xiao \bgroup et al.\egroup }2023]{xiao2023bge}
Shitao Xiao, Zheng Liu, Peitian Zhang, Niklas Muennighoff, Defu Lian, and Jian-Yun Nie.
\newblock 2023.
\newblock C-pack: Packed resources for general chinese embeddings.
\newblock {\em arXiv preprint arXiv:2309.07597}.

\bibitem[\protect\citename{Yin and Zubiaga}2021]{yin2021towards}
Wenjie Yin and Arkaitz Zubiaga.
\newblock 2021.
\newblock Towards generalisable hate speech detection: a review on obstacles and solutions.
\newblock {\em PeerJ Computer Science}, 7:e598.

\bibitem[\protect\citename{Zhang \bgroup et al.\egroup }2023]{zhang2023twostage}
Longhui Zhang, Yanzhao Zhang, Dingkun Long, Pengjun Xie, Meishan Zhang, and Min Zhang.
\newblock 2023.
\newblock A two-stage adaptation of large language models for text ranking.
\newblock {\em arXiv preprint arXiv:2311.16720}.

\bibitem[\protect\citename{Zheng \bgroup et al.\egroup }2024]{zheng2024llamafactory}
Yaowei Zheng, Richong Zhang, Junhao Zhang, Yanhan Ye, and Zheyan Luo.
\newblock 2024.
\newblock Llama-factory: Unified efficient fine-tuning of 100+ language models.
\newblock In {\em Proceedings of the 62nd Annual Meeting of the Association for Computational Linguistics (Volume 3: System Demonstrations)}, pages 400--410.

\end{thebibliography}

\end{document}